\documentclass[10pt, onecolumn]{article}

\usepackage{scrextend}
\usepackage{graphicx}
\usepackage{color}
\usepackage{multirow}
\usepackage{authblk}

\providecommand{\keywords}[1]{\textbf{\textit{Index terms---}} #1}
\newcommand{\quotes}[1]{``#1''}
\makeatletter
\newcommand{\specificthanks}[1]{\@fnsymbol{#1}}
\makeatother

\graphicspath{{./}} 
\begin{document}

\title{Deep-Learning
Convolutional Neural Networks for scattered shrub detection with Google
Earth Imagery}

\author[1]{Emilio Guirado\footnote{Both authors have contributed equally to this work}}
\author[2]{Siham Tabik$^*$}
\author[1,3,5]{Domingo Alcaraz-Segura}
\author[2,4]{Javier Cabello}
\author[2]{Francisco~Herrera}
\affil[1]{Andalusian Center for Assessment and monitoring of global
change (CAESCG), Univ. of Almer\'ia, Almer\'ia 04120, Spain}
\affil[2]{Soft Computing and Intelligent Information Systems
research group, Univ. of Granada, 18071 Granada,
Spain}
\affil[3]{Dept. Botany, Univ. of
Granada, Granada, 18071, Spain}
\affil[4]{Dept. Biology and Geology, Univ. of Almer\'ia, 04120, Spain \\
 emails: e.guirado@ual.es, siham@ugr.es}

\maketitle

\begin{abstract}
There is a growing demand for accurate high-resolution land cover
maps in many fields,  e.g., in land-use planning and biodiversity
conservation. Developing such maps has been performed using
Object-Based Image Analysis (OBIA) methods, which usually reach good
accuracies, but require a high human supervision and the best
configuration for one image can hardly be extrapolated to a
different image. Recently, the deep learning Convolutional Neural
Networks (CNNs) have shown outstanding results in object recognition
in the field of computer vision. However, they have not been fully
explored yet in land cover mapping  for detecting species of high
biodiversity conservation interest. This paper analyzes the
potential of CNNs-based methods for plant species detection using
free high-resolution Google Earth$^{TM}$ images and provides an
objective comparison with the state-of-the-art OBIA-methods. We
consider as case study the detection of {\it Ziziphus lotus} shrubs,
which are protected as a priority habitat under the European Union
Habitats Directive. According to our results, compared to OBIA-based
methods, the proposed CNN-based detection model, in combination with
data-augmentation, transfer learning and pre-processing, achieves
higher performance with less human intervention and the knowledge it
acquires in the first image can be transferred to other images,
which makes the detection process very fast. The provided
methodology can be systematically reproduced for other species
detection.
\end{abstract}


\section{Introduction}

Changes in land cover and land use are pervasive, rapid, and can
have significant impact on humans, the economy, and the environment.
Accurate land cover mapping is of paramount
importance in many applications, e.g., in urban planning, forestry,
 natural hazard mapping, habitat mapping, assessment of land-use
 change effects on climate, etc.~\cite{congalton2014global,rogan2004remote}.

In practice, land cover maps are built by analyzing remotely sensed
imagery, captured by satellites, airplanes or drones, using
different classification methods. The accuracy of the results and
their interpretation depends on the quality of the input data, e.g.,
spatial, spectral, and radiometric resolution of the images, and
also on the classification methods used. The most used methods can
be divided into two categories: pixel-based classifiers and
Object-Based Image Analysis (OBIA)~\cite{blaschke2010object}.
Pixel-based methods, which use only the spectral information
available for each pixel, are faster but ineffective especially for
high resolution images~\cite{li2014object,pierce2015accuracy}.
Object-based methods take into account the spectral as well as the
spatial properties of image objects, i.e., set of neighbor similar
pixels.

OBIA methods are more
accurate but very expensive from a computational point of view, they
require a high human supervision and number of iterations to obtain acceptable
accuracies, and are not easily portable to other images (e.g.,
to other areas, seasons, extensions, radiometric calibrations or different
spatial or spectral resolutions). To detect a specific
object in an input image, first, the OBIA method segments
the image (e.g., by using a multi-resolution segmentation
algorithm), and then classifies the segments based on their similarities (e.g., by
using algorithms such as the k-nearest neighbor). This procedure has
to be repeated for each single input image and  the knowledge acquired
from one input image cannot be reutilized in another.

In the last five years, deep learning and particularly supervised
Convolutional Networks (CNNs) based models have demonstrated
impressive accuracies in object recognition and image classification
in the field of computer
vision~\cite{krizhevsky2012imagenet,Le2013,Hinton2012,Sainath2013}.
This success is due to the availability of larger datasets, better
algorithms, improved network architectures,  faster GPUs and also
improvement techniques such as,  transfer-learning and data-augmentation.

This paper analyzes the potential of CNNs-based methods for plant
species mapping using high-resolution Google Earth$^{TM}$ images and
provides an objective comparison with the state-of-the-art
OBIA-based methods.

As case study, this paper addresses the challenging problem of
detecting {\it Ziziphus lotus} shrubs, a species known by its role
as the dominant plant that characterizes an ecosystem of priority
conservation in the European Union \quotes{Arborescent matorral}
with {\it Ziziphus}, which is experiencing a serious decline during
the last decades. The complexity of this case is due to the fact
that {\it Ziziphus lotus} individuals are scattered arborescent
shrubs with variable shapes, sizes, and distribution patterns. In
addition, distinguishing {\it Ziziphus lotus} shrubs from other
neighbor plants in remote sensing images is complex for non-experts
and for automatic classification methods.

From our results, compared to OBIA, the detection model based on
GoogLeNet network, in combination with data-augmentation,
transfer-learning (fine-tuning) and pre-processing the input test
images, achieves higher precision and balance between recall and
precision in the problem of {\it Ziziphus lotus} detection. In
addition, the detection process using GoogLeNet detector is faster,
which implies a high user productivity in comparison with OBIA.

In particular, the contributions of this work are:
\begin{itemize}
\item Developing an accurate CNN-based detection model for plant individuals
mapping using high-resolution remote sensing images, extracted from
Google Earth$^{TM}$.
\item Designing a new dataset containing images of Ziziphus lotus individuals and bare
soil with sparse vegetation for training the CNNs-based  model.
\item Demonstrating that the use of small  datasets to train GoogLeNet-model with transfer learning
 from ImageNet (i.e., fine-tuning) can lead to satisfactory results that
can be further enhanced by including data-augmentation, and specific
pre-processing techniques.
\item Comparing a CNN-based model with an OBIA-based method in terms of performance, user productivity,
and transferability to other regions.
\item Providing a complete description of the used methodology so that it can be reproduced by other
researchers for the classification and detection of other plant
species.
\end{itemize}

This paper is organized as follows. A review of related works is
provided in Section~2. A description of the proposed CNN-methodology
is given in Section~3. The considered study areas and how the
dataset were constructed to train the CNN-based classifier can be
found in Section~4. The experimental results of the detection using
CNNs and OBIA are provided in Section~5 and finally conclusions.

\section{Related Works}
This section reviews the related work in land cover mapping then
explains how OBIA, the state-of-the-art methods are used for plant
species detection.
\subsection{Land cover mapping}
In the field of remote sensing, object detection has been
traditionally performed using pixel-based classifiers or
object-based methods~\cite{pierce2015accuracy}. Several papers
have demonstrated that Object-Based Image Analysis (OBIA) methods
are more accurate than pixel based methods, particularly for high
spatial resolution images~\cite{blaschke2010object}. In the field
of computer vision, object detection within an image is more
challenging than scene tagging or classification because it is
necessary to determine the image segment that contains the searched
object. In most object detection works, first a classifier is
trained and then it is
run either on a number of sliding windwos or on the segmented input
image.

Recently, deep learning CNNs have started to be used for
scene tagging and object detection in remotely sensed
images~\cite{zhao2017object,langkvist2016classification,hu2015transferring,santara2016bass}.
However, as far as we know, there does not exist any study on using CNNs in plant species detection or comparison between
OBIA and deep CNNs methods.

The existing works that use deep CNNs in remotely sensed images can
be divided into two broad groups. The first group focuses on the
object detection or classification of high-resolution multi-band
imagery, i.e., with a spectral dimension greater than three, and
apply CNNs-based methods at the pixel level, i.e., using only the
spectral
information~\cite{zhao2017object,langkvist2016classification}. The
second group focuses on the classification or tagging of whole
aerial RGB images, commonly called scene classification, and show
their accuracies using bechmark databases such as, UC-Merced
dataset~\footnote{http://vision.ucmerced.edu/datasets/landuse.html}
and Brazilian Coffee Scenes
dataset~\footnote{www.patreo.dcc.ufmg.br/downloads/brazilian-coffee-dataset/}~\cite{castelluccio2015land,hu2015transferring}.
It is worth to mention that these datasets contain a large number of
manually labeled images. For example, the Brazilian Coffee Scenes
dataset contains $50,000$ of $64\times64$-pixel tiles, labeled as
coffee (1,438) non-coffee (36,577) or mixed (12,989) and UC-Merced
dataset contains 2100 $256\times256$-pixel images labeled as
belonging to 21 land use classes, 100 images corresponding to each
class. Several works have reached classification accuracies greater
than $95\%$ on these database
~\cite{castelluccio2015land,hu2015transferring}. 

The most related work to ours is~\cite{li2016deep}. It addresses
the detection of palm oil trees in agricultural areas using four
bands imagery with $0.5\times0.5$m spatial resolution via CNNs.
Since oil-palm trees have the same age, shape, size, and are placed
at the same distance from each other, the authors could combine
LeNet-based classifier with a very simple detecting technique. In
addition, the authors used a large number of manually labeled
training samples, 5000 palm tree samples and 4000 background
samples.

Our study is more challenging, because {\it Ziziphus lotus}
is not a crop, it is a wild plant that has very different shapes,
sizes, and intensities of the green color. In addition, we will show
in this paper that a smaller training set can also lead to
competitive results.
\begin{figure}[h!]
\centering
\includegraphics[width=1.0\textwidth]{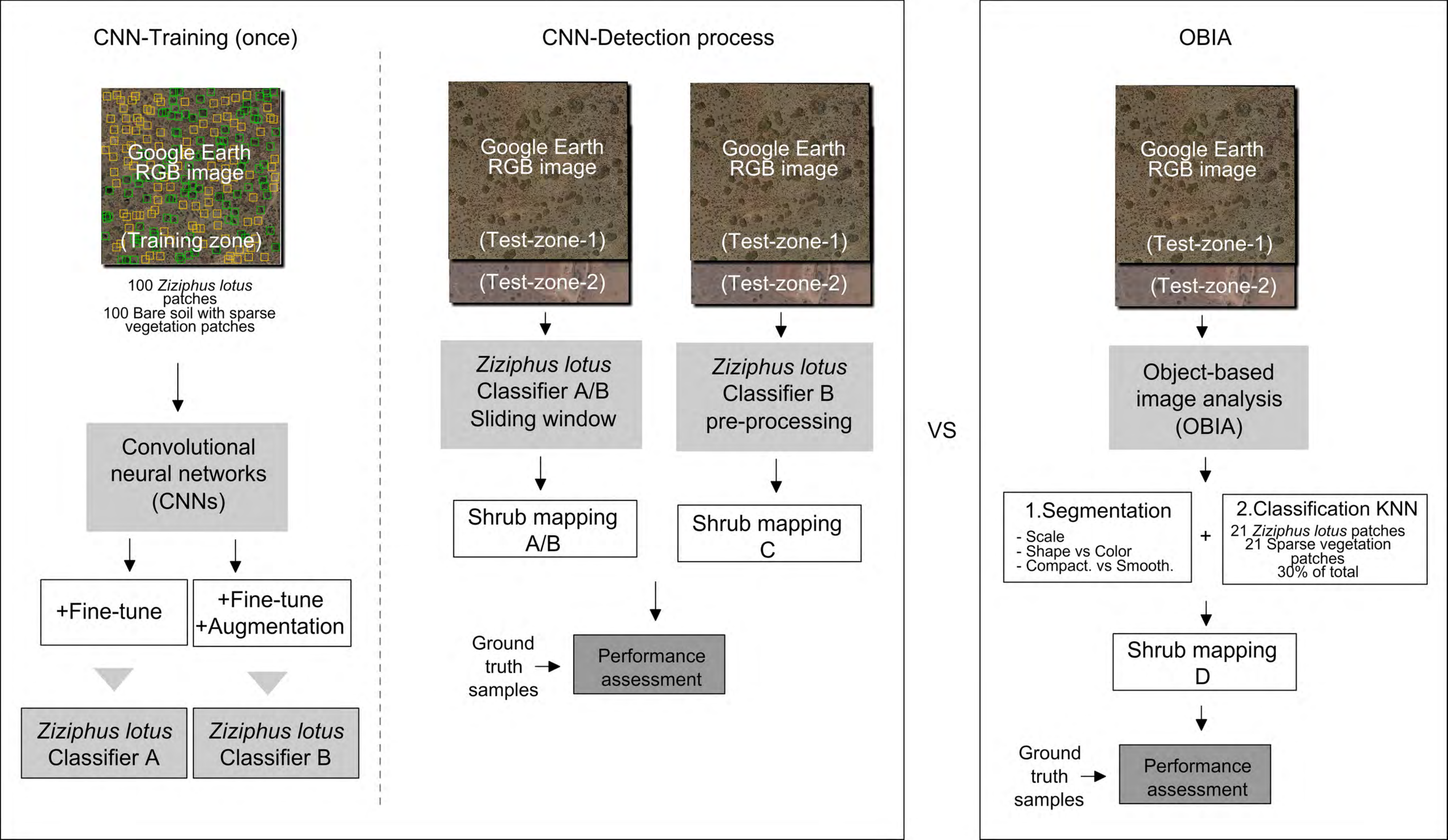}\\
~~~~~~~~~~~~(a) CNN~~~~~~~~~~~~~~~~~~~~~~~~~~(b) OBIA
\caption{Flowchart of the {\it Ziziphus lotus} shrub mapping process
using (a) Convolutional neural networks (CNNs), considering two detection approaches: slidning window and pre-processing and (b) Object-based
image analysis (OBIA).}
 \label{workflow}
\end{figure}

\subsection{OBIA-based detection}
Differently to CNNs-based approach, OBIA-based detector does not
re-utilize the learning from one image to another. The detection is
applied from scratch on each individual image. The OBIA detection
approach is performed in two steps. First, the input image is
segmented, and then each segment is analyzed by a classification
algorithm. A simplistic flowchart of the CNNs- and OBIA-based
approaches is illustrated in Figure~\ref{workflow}(b).
The OBIA-detector used in this study is
implemented in E-cognition 8.9 software~\cite{ecognition} and
works in two steps as follows:


\begin{itemize}
\item Segmentation step: first, the input image is segmented using the
multi-resolution algorithm~\cite{baatz2000multiresolution}. In this step,
 the user has to manually
initialize a set of non-dimensional parameters namely: i)
 The scale parameter, to define the maximum standard deviation of
 the homogeneity criteria (in color and shape) in regard to the weighted images layers for
 resulting image objects. The higher the value, the larger the
 resulting image objects. ii) The shape versus color
 parameter, to prioritize homogeneity in color versus in shape or texture when creating
 the image objects. Values closer to one indicate shape priority,
 while values closer to zero indicate color priority. iii) The
 compactness versus smoothness parameter, to prioritize whether producing
 compact objects over smooth edges during the segmentation. Values
 closer to one indicate compactness priority,
while values closer to zero indicate smoothness
priority~\cite{tian2007optimization}.

\item Classification step: the resulting segments are classified using the
K-Nearest Neighbor (KNN) method. For this, the user identifies
sample sites for each class and calculates their statistics. Then, objects are classified based on their
resemblance to the training sites using the calculated statistics.
Finally, the validation of the classification is carried out using an
independent set of field samples. KNN typically uses 30\% of labeled
field samples for training, i.e. calculating the statistics, and
70\% of the field samples to evaluate the classifier. It provides a
confusion matrix to calculate the commission and omission errors,
and the overall accuracy~\cite{congalton2008assessing}.

\end{itemize}

\section{CNN-based detection for plant species mapping}

We reformulate the problem of detecting a particular plant
species into a two-class problem, where the true class is
\quotes{{\it Ziziphus lotus} shrubs}  and the false class is
\quotes{bare soil with
sparse vegetation}. 
To build the CNNs-based detection model, we first designed a
field-validated training dataset, then we built a classification
model (CNN classifier), and finally, during the detection process, we considered two solutions, the
sliding-window technique and a set of pre-processing techniques to localize {\it Ziziphus lotus}
in the test scenes. A simplistic flowchart of the CNNs- and OBIA-based approaches is
illustrated in Figure~\ref{workflow}(a).


\subsection{Classification model using fine-tuning}
\begin{figure}[h!]
\centering
\includegraphics[width=1.0\textwidth]{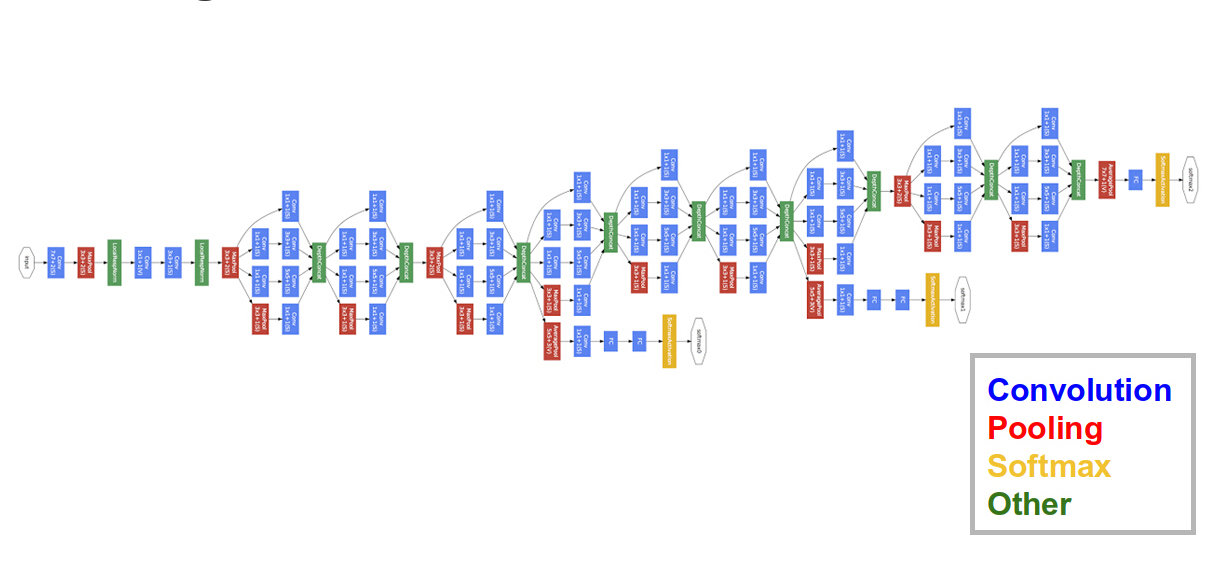}
\caption{Architecture of GoogLeNet
network~\cite{szegedy2015going}. } \label{googARCH}
\end{figure}

In this work, we use feed-forward Convolutional Neural Networks
(CNNs) for supervised classification, as they have provided very
good accuracies in several applications. These methods automatically
discover increasingly higher level features from
data~\cite{krizhevsky2012imagenet,guo2016deep}. The lower
convolutional layers capture low-level image features, e.g. edges,
color, while higher convolutional layers capture more complex
features, i.e., composite of several features.

We use GoogLeNet model~\cite{szegedy2015going}, which was the
winner of ILSVRC (ImageNet Large Scale Visual Recognition
Competition (ILSVRC)) 2014. Compared with previous network
architectures, GoogLeNet provides higher accuracy with less
computational cost. It has 12$\times$ fewer parameters than AlexNet,
the network that won ILSVRC 2012. GoogLeNet has 6.8 million
parameters and 22 layers with learnable weights organized in four
parts: i) the initial segment, made up of three convolutional
layers, ii) nine inception modules, each module is a set of
convolutional and pooling layers at different scales performed in
parallel then concatenated together, iii) two auxiliary classifiers,
each classifier is actually a smaller convolutional network put on
the top of the output of an intermediate inception module, and iv)
one output classifier. See Figure~\ref{googARCH}.

Deep CNNs, such as Googlenet, are generally trained based on the
prediction loss minimization. Let $x$ and $y$ be the input images
and corresponding output class labels, the objective of the training
is to iteratively minimize the average loss defined as

\begin{equation}
J ( w ) =\frac{1}{N}  \sum_{i=1}^N L ( f ( w ; x_i ), y_i ) +
\lambda R ( w )
\end{equation}

\noindent This loss function measures  how different is the output
of the final layer  from the ground truth. $N$ is the number of data
instances (mini-batch) in every iteration, $L$ is the loss function,
$f$ is the predicted output of the network depending on the current
weights $w$, and R is the weight decay with the Lagrange multiplier
$\lambda$. It is worth to mention that in the case of GoogLeNet, the
losses of the two auxiliary classifiers are weighted by $0.3$ and
added to the total loss of each training iteration. The Stochastic
Gradient Descent (SGD) is commonly used to update the weights.

\begin{equation}
w_{t +1} = \mu w_t - \alpha  \Delta J ( w_t )
\end{equation}

\noindent where $\mu$ is the momentum weight for the current weights
$w_t$ and $\alpha$ is the learning rate.

The network weights, $w_t$, can be randomly initialized if the
network is trained from scratch. However, this is suitable only when
a large labeled training-set is available, which is expensive in
practice. Several works have shown that
data-augmentation~\cite{tabik2017snapshot} and transfer
learning~\cite{shin2016deep} help overcoming this limitation.

\begin{itemize}
\item {\bf Transfer learning} (e.g. fine-tuning in CNNs) consists of
re-utilizing the knowledge
learnt from one problem to another related
one~(\cite{pan2010survey}). Applying transfer learning with deep
CNNs depends on the similarities between the original and new
problem and also on the size of the new training set. In deep CNNs,
transfer learning can be applied via fine-tuning, which involves
initializing the weights of the network by the pre-trained weights
on a different dataset.

In general, fine-tuning the entire network, i.e., updating all the
weights, is only used when the new dataset is large enough,
otherwise, the model could suffer overfitting especially among the
first layers of the network. Since these layers extract low-level
features, e.g., edges and color, they do not change significantly
and can be utilized for several visual recognition tasks. The last
learnable layers of the CNN are gradually adjusted to the
particularities of the problem and extract high level features.

In this work, we have used fine-tuning GoogleNet and initialized it
with the pre-trained weights of the same architecture on ImageNet
dataset (around 1.28 million images over 1,000 generic object
classes)~\cite{krizhevsky2012imagenet}.

\item {\bf Data-augmentation}, also called
data transformation or distortion, it is used to increase the volume of
the training set by applying specific deformations on the input
images, e.g., rotation, translation. The set of transformations that
improves the performance of the CNN-model depends on the
particularities of the problem.

\end{itemize}

\subsection{Plant species detection}
To obtain an accurate detection in a new image, different from the
image used for training the CNN-classifier, we analyzed two
approaches:
\begin{itemize}
\item {\bf Sliding window} is a technique frequently used for
detection. The detection task consists of applying the obtained
GoogLeNet-classifier at all locations and scales of the input image.
The sliding window approach is an exhaustive method since it
considers a very large number of candidate windows of different
sizes and shapes across the input image. The classifier is then run on each one of
these windows. To maximize the detection accuracy, the probabilities
obtained from different window sizes can be assembled into one
heatmap. Finally, probability heatmaps are usually transformed into
classes using a thresholding technique, i.e., areas with
probabilities higher than 50\% are usually classified as the true
class (e.g. {\it Ziziphus lotus}) and areas with probabilities lower
than 50\% as background (e.g. bare soil with sparse vegetation).

\item {\bf Pre-processing} techniques can also help to improve the
detection accuracy and execution time. The set of pre-processing
techniques that provides the best results depends on the nature of
the problem and the object of interest. From the multiple techniques
that we explored, the ones that provided the best detection performance were: i) Eliminating the
background using a threshold based on its typical color or darkness
(e.g. by converting the RGB image to gray scale, grays lighter than
100 digital level corresponded to bare ground).  ii) Applying an
edge-detection method that filters out the objects with an area or
perimeter smaller than the minimum size of the target objects (e.g.
180  pixels of resolution $0.5\times0.5$ m, correspond to the area
of the smallest {\it Ziziphus lotus} individual in the image, around
22 m$^2$).
\end{itemize}

\section{Study areas and datasets construction}

This Section describes the study areas and provides full details on
how the training and test sets were built using Google Earth$^{TM}$
images. We consider the challenging problem of detecting {\it
Ziziphus lotus} shrubs, since this is the key species of an
ecosystem of priority conservation in the European Union (habitat
5220* habitat of 92/43/EEC Directive) that is declining during the
last decades in SE Spain, Sicily, and Cyprus, where the only
European populations occur (\cite{tirado2003shrub}). In Europe, the
largest population occurs in the Cabo de Gata-N\'ijar Natural Park
(SE Spain), where an increased mortality of individuals of all ages
has been observed in the last decade~(\cite{guirado2015factores}).

\subsection{Study areas}
In this study, we considered three zones: one training-zone, for
training the CNN-model, and two test zones (labeled as test-zone-1
and test-zone-2) for testing and comparing the performance of both
CNN- and OBIA-based models.
\begin{figure}[h!]
\centering
\includegraphics[width=1.0\textwidth]{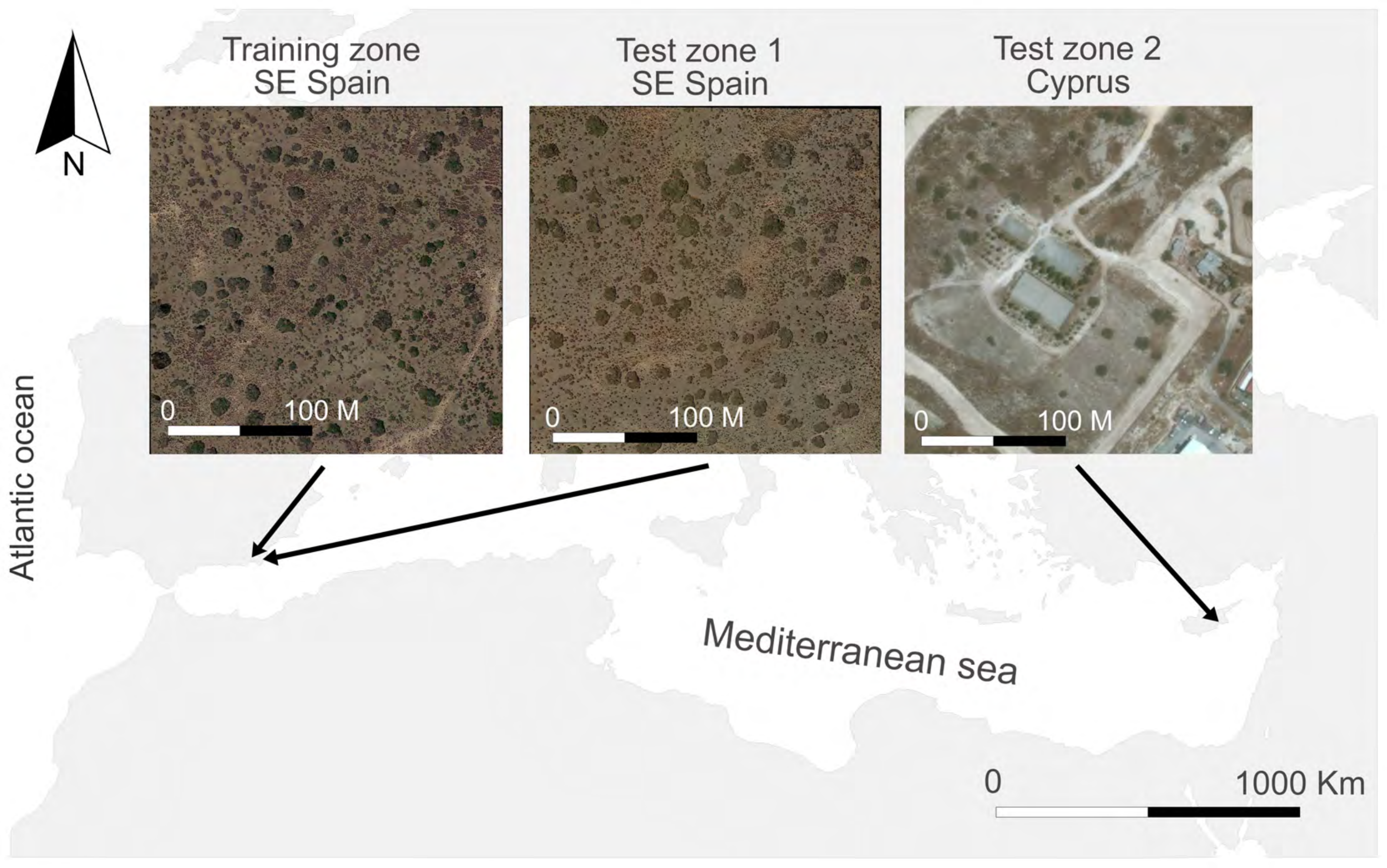}
\caption{Localization of the three study areas used in this work:
Training-zone and Test-zone-1 in Cabo de Gata-N\'ijar Natural Park
(Spain), and Test-zone-2 in Rizoelia National Forest Park (Cyprus).
The three images are $230\times230$ m with a native resolution in
Google Earth$^{TM}$ of 0.5 m per pixel (but downloaded as
$1900\times1900$ pixel images). {\it Ziziphus lotus} shrubs can be
seen in the three images. The used projection was geographic with
the WGS84 Datum.} \label{zoneA_zoneB}
\end{figure}

\begin{itemize}
\item {\bf The training-zone} used for training the CNN-based model.
This zone is located in Cabo de Gata-N\'ijar Natural Park,
36$^{\circ}$49'43'' N, 2$^{\circ}$17'30'' W, in the province of
Almer\'ia, Spain (Figure~\ref{zoneA_zoneB}). The climate is
semi-arid Mediterranean. The vegetation is scarce and patchy, mainly
dominated by large {\it Ziziphus lotus} shrubs surrounded by a
heterogeneous matrix of bare soil and small scrubs (e.g. {\it Thymus
hyemalis}, {\it Launea arborescens} and {\it Lygeum spartum}) with
low coverage~(\cite{tirado2003shrub,rivas1944formaciones}). {\it
Ziziphus lotus} forms large hemispherical bushes with very deep
roots and 1-3 m tall that trap and accumulate sand and organic
matter building geomorphological structures, called nebkhas, that
constitute a shelter micro-habitat for many plant and animal
species~(\cite{tirado2003shrub,tirado20095220,lagarde2012bushes}).

\item {\bf Test-zone-1} and {\bf  test-zone-2} belong to two different protected
areas. Test-zone-1 is located $1.5$ km west from the training-zone,
36$^{\circ}$49'45'' N, 2$^{\circ}$17'30'' W. Test-zone-2 is located in
Rizoelia National Forest Park in Cyprus, 34$^{\circ}$56'08'' N,
33$^{\circ}$34'27'' (Figure~\ref{zoneA_zoneB}). These two test-zones
are used for comparing the performance between CNNs and OBIA for
detecting {\it Ziziphus lotus}.
\end{itemize}

\subsection{Datasets construction}
In OBIA, the training dataset consists of a set of georreferenced
points from each one of the classes that we want to classify. These
points must fall in the same scene that we aim to classify.
Conversely, for CNNs, the training dataset consists of a set of
images that contain the object of interest, but these images do not have
to belong to same scene that we aim to classify, allowing for
transferability to other regions, which is an advantage of CNNs over
the OBIA method.

The satellite RGB orthoimages used were downloaded from Google
Earth$^{TM}$ in WGS84 Mercator (sphere radius 6378137 m, European
Petroleum Survey Group (EPSG) 3785). The scenes of the three areas,
training-zone, test-zone-1 and test-zone-2, are of size
$230\times230$ meters. The images were downloaded using the maximum zoom
level in their area. The obtained scenes have $1900\times1900$
pixels, with a pixel size of 0.12 m due to the smoothing applied by
Google Earth at the highest zoom, however the real resolution is approximately 0.5 m. The capturing date was
June 22th 2016 for the two zones in Spain, and June 20th 2015 for
Cyprus.

We addressed the {\it Ziziphus lotus} detection problem by
considering two classes, 1) {\it Ziziphus lotus} shrubs class and 2)
Bare soil with sparse vegetation class.

\subsubsection{Datasets for training OBIA and for ground truthing}

\begin{itemize}
\item  In test-zone-1, from Spain, 72 {\it Ziziphus lotus} individuals were identified
in the field. The perimeter of each was georeferenced in the field
with a differential GPS, GS20, Leica Geosystems, Inc. From the 72
individuals, 30\% (21 individuals) were used for training the OBIA
method and 70\% (51 individuals) as ground truth both for the OBIA
and the CNN method.

\item In test-zone-2, from Cyprus, 41 {\it Ziziphus lotus} individuals were visually
identified in Google Earth by the authors using descriptions
provided by local experts (\cite{browning2009field}). From the 41
individuals, 30\% (12 individuals) were used for training the OBIA
method and 70\% (29 individuals) as ground truth both for the OBIA
and the CNN method. In both test zones, the corresponding number of
points (72 and 41, respectively) were georeferenced for the Bare
soil with sparse vegetation class.
\end{itemize}

\subsubsection{Training dataset for the CNN-classifier}

\begin{figure}[h!]
\centering
\includegraphics[width=0.6\textwidth]{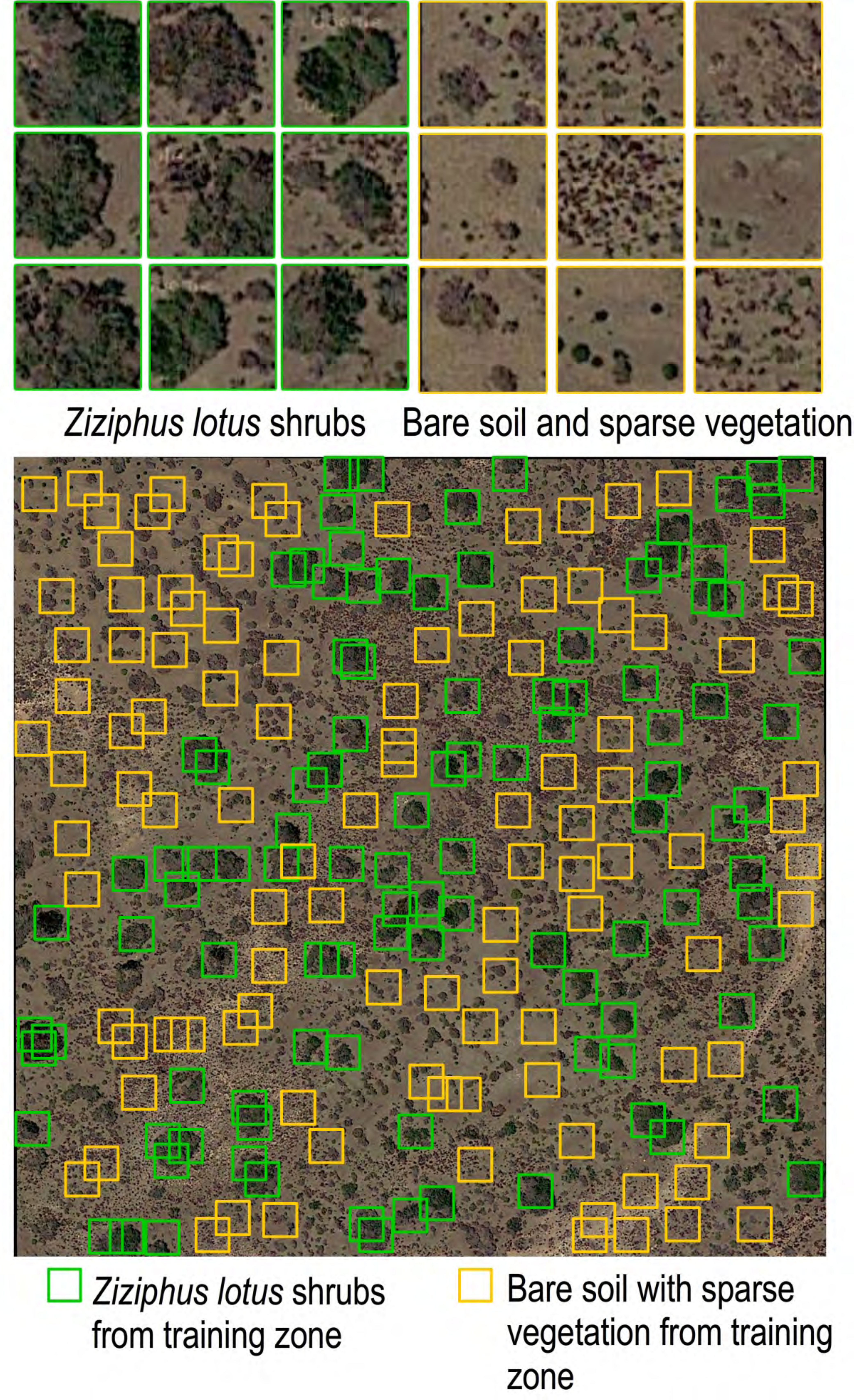}
\caption{The two top panels show examples of the $80\times80$-pixels
image patches used to build the training dataset for the CNN model: left)
patches of {\it Ziziphus lotus} class, right) patches of Bare soil with sparse
vegetation class. The bottom panel shows the training-zone dataset with
{\it Ziziphus lotus} patches labeled with a
green contour and Bare soil and sparse vegetation patches
labeled with yellow contour.} \label{fig:gt}
\end{figure}

The design of the training dataset is key to the performance of a
good CNN classification model. For labeling the training dataset, we
identified 100 $80\times80$-pixels image patches containing {\it
Ziziphus lotus} shrubs and 100 images for Bare soil with sparse
vegetation. Examples of the labeled classes can be seen in
Figure~\ref{fig:gt}. We distributed the 100 images of each class
into 80 images for training and 20 images for validating the
obtained CNNs classifiers, as summarized in Table~\ref{tab1}.

\begin{table}[h!]
\caption{Training and testing datasets for both CNN (Convolutional
Neural Networks) and OBIA (Object-Based Image Analysis) used for
mapping {\it Ziziphus lotus} shrubs. Bare soil: Bare soil and sparse
vegetation; Img: 80x80-pixel images; Poly: digitized polygons}
\begin{center}
    \begin{tabular}{|l|l|l||l|l||l|l|}
    \hline
Class    & \multicolumn{2}{|l|}{CNN Classifier} & \multicolumn{2}{||l||}{OBIA classifier} & \multicolumn{2}{|l|}{Accuracy}
\\\hline
         & training      &  validation & \multicolumn{2}{|l||}{training} & \multicolumn{2}{|l|}{assessment} \\ \hline\hline
         & \multicolumn{2}{|l||}{ {\footnotesize Training-zone}} & {\footnotesize Test-zone-1}& {\footnotesize Test-zone-2}       & {\footnotesize Test-zone-1}    & {\footnotesize Test-zone-2}    \\ \hline
{\it Ziziphus}&  {\footnotesize 80 img} &  {\footnotesize 20 img} &  {\footnotesize 21 poly}  &  {\footnotesize 15 poly} &  {\footnotesize 51 poly} &  {\footnotesize 36
poly}\\ \hline Bare soil   &  {\footnotesize 80 img}   &  {\footnotesize 20 img}  &  {\footnotesize 21 poly} &  {\footnotesize 15 poly}
&  {\footnotesize 51 poly} &  {\footnotesize 36 poly} \\\hline
   \end{tabular}
\end{center}
\label{tab1}
\end{table}

\section{Experimental evaluation and accuracy assessment}

This section is organized in two parts. The first part describes the
steps taken to improve the baseline detection results of the
CNN-based model (GoogLeNet). In particular, we considered transfer-learning
(fine-tuning) and data-augmentation to improve the classifier
training, and selecting the best sliding-window size and
pre-processing (background elimination and long-edge detection) to
improve the classifier detection performance. The second part
describes the classification steps used with OBIA and provides a
comparison between GoogeLeNet- and OBIA-models. For the evaluation
and comparison of accuracies, we used three metrics, {\it precision}
(also called positive predictive value, i.e., how many detected
Ziziphus are true), {\it recall} (also known as sensitivity, i.e.,
how many actual Ziziphus were detected), and {\it F1 measure}, which
evaluates the balance between {\it precision} and {\it recall}. Where
$$precision=\frac{True~Positives}{True~Positives+False~Positives},$$
$$recall=\frac{True~Positives}{True~Positives+False~Negatives},$$ and
$$F1~measure=2\times\frac{precision \times recall}{precision
+recall}$$

\subsection{CNN training with fine-tuning and augmentation}

\begin{figure}[h!]
\centering
\includegraphics[width=0.46\textwidth]{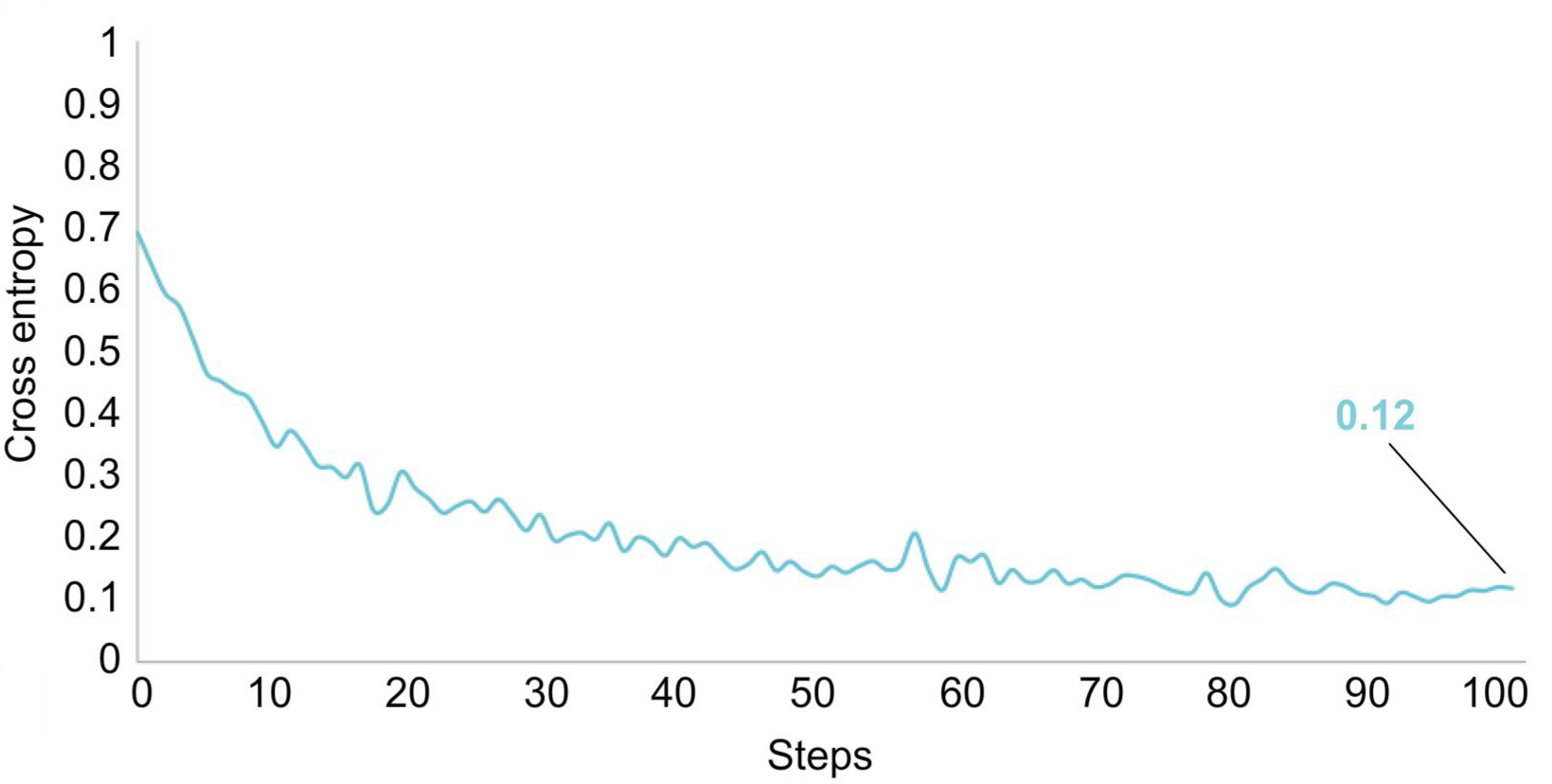}
\includegraphics[width=0.46\textwidth]{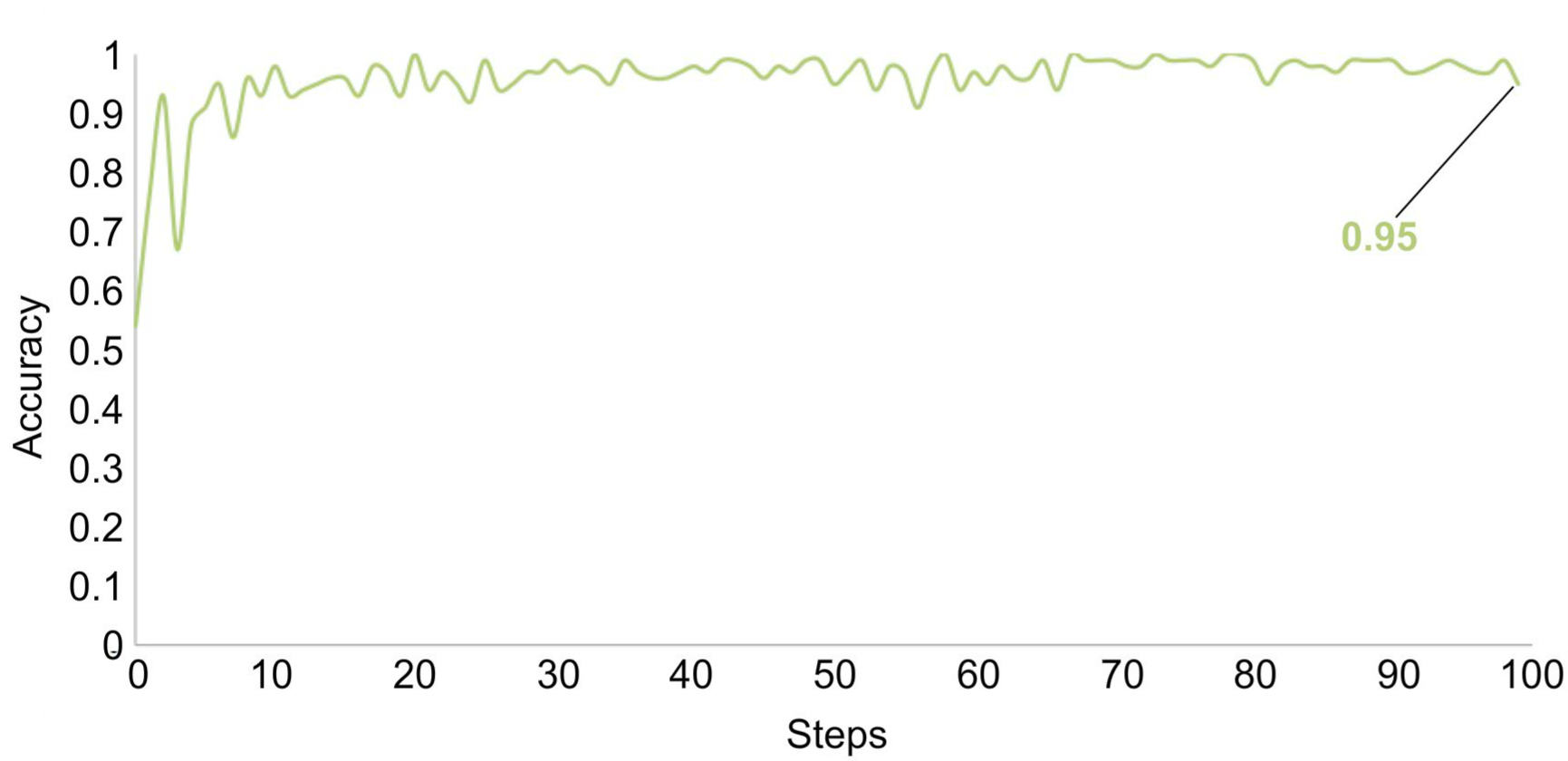}
\caption{The  accuracy and cross entropy of the GoogLeNet-classifier with fine-tuning
and data-augmentation during the training
process of {\it Ziziphus lotus} shrubs with Google Earth images.}
\label{convergence}
\end{figure}

For the experiments with GoogleNet-based model, we have used the
open source software library
Tensorflow~(\cite{abadi2016tensorflow}). 
To improve the accuracy
and reduce overfitting we i) used fine-tuning by initializing the
model with the pre-trained weights of ImageNet, and ii) applied data
augmentation techniques to increase the size of the dataset from 100
to 6000 images. In particular, we applied:
\begin{itemize}
\item Random scale: increases the scale of the image by a factor picked randomly in [1 to 10\%]
\item Random crop: crops the image edges by a margin in [0 to 10\%]
\item Flip horizontally: randomly  mirrors the image from left to right.
\item Random brightness: multiplies the brightness of the image by a factor picked randomly in [0, 10].
\end{itemize}

The obtained classifier converges within the first 100 training
iterations as shown in Figure~\ref{convergence}.

\begin{figure}[h!]
\centering
\includegraphics[width=0.85\textwidth]{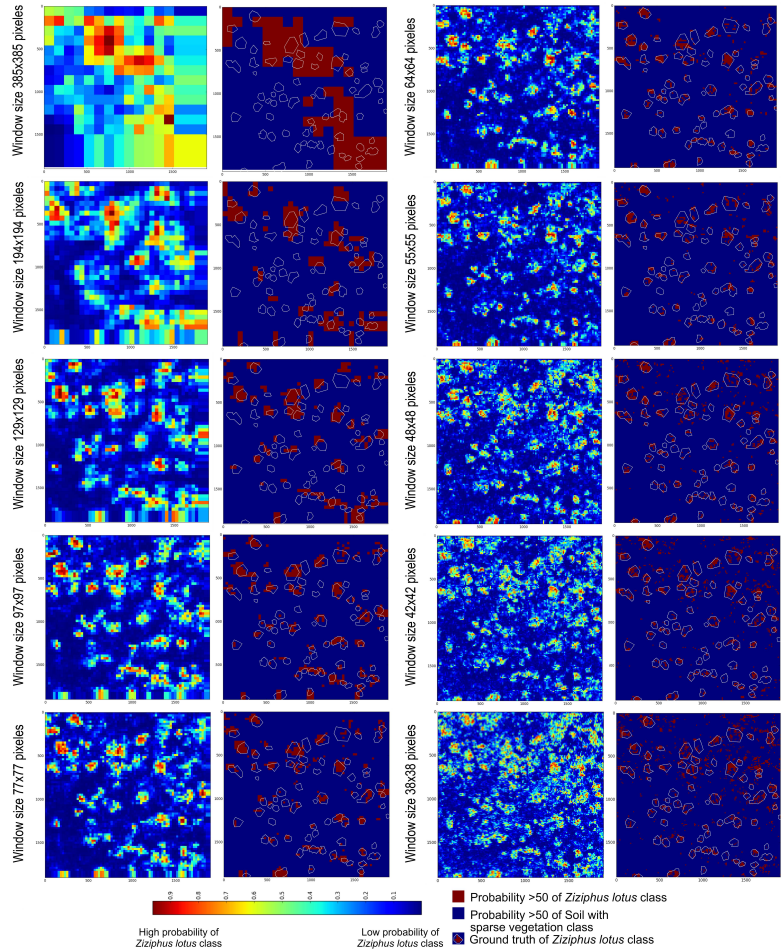}
\caption{ Maps showing the probability of {\it Ziziphus lotus}
presence according to the GoogLeNet-classifier trained with
fine-tuning and data-augmentation and applied on different
sliding-window sizes from $38\times 38$ to $385\times 385$ pixels in
Test-zone-1. The first and third columns show the heatmaps of the
probability of {\it Ziziphus lotus} presence, and the second and
fourth columns show the corresponding binary maps after applying a
threshold of probability greater than 50\%. The white polygons
correspond to the ground-truth perimeter of each individual
georeferenced in the field with a differential GPS.}
\label{heatmaps}
\end{figure}

\subsubsection{Detection with Sliding windows}
\begin{table}[h!]
  \centering
   \caption{The CNN-detection results in Test-zone-1 with different sliding window sizes.
   Accuracies are expressed in terms of true positives (TP), false positives (FP) and false negatives (FN), precision, recall, F1-measure, and execution time of the detection process.}
   \begin{tabular}{|l|l|l|l|l|l|l|l|l|}
   \hline
win. size& total \# of   &TP &FP &FN &precis. &recall &F1- &time \\
 (pixels) & win.  & & & &(\%) &(\%) & meas.(\%)&(min)\\\hline
\hline
385$\times$385  &196   & 31 &18 &41 &63.27  &43.06  & 51.24 & 6.0\\
194$\times$194  &961   & 34 &7  &38 &  82.93 & 47.22 &60.18 & 29.4\\
129$\times$129  &2209  & 42 &6  &30 & 87.50 &58.33 &70.00 & 67.6\\
97$\times$ 97   &4096  & 59 &6  &13 &  90.77 &81.94 &             86.13& 125.4\\
77$\times$77    &5929  & 59 &5  &13 & {\bf  92.19} &81.94 &86.76& 181.5\\
64$\times$64    &9506  & 65 &7  &7&  90.28  &{\bf 90.28} &{\bf 90.28}&  291.0\\
55$\times$55    &13340 & 65 &12 &7&  84.42 &90.28 &87.25 & 408.4\\
48$\times$48    &17292 & 68 &16 &4&  80.95 &94.44 &87.18&  529.3\\
42$\times$42    &22200 & 70 &17 &2&  80.46 &97.22 &88.05 & 679.6\\
38$\times$ 38   &27888 & 71 &39 &1 &  64.55 & 98.61&78.02 &
853.7\\\hline
   \end{tabular}
   \label{swindow}
\end{table}

To assess the ability of CNNs model to detect {\it Ziziphus lotus}
shrubs in Google Earth images, we applied the trained GoogLeNet
classifier across the entire scene of test-zone-1 by using the
sliding window technique. Since the diameter of the smallest {\it Ziziphus lotus}
individual georeferenced in the field was 4.6 m ($38$ pixels) and the largest
individual in the region has a diameter of 47 m ($385$ pixels), we evaluated a
range of window sizes
between $38\times38$ and $385\times385$ pixels and a horizontal and
vertical sliding step of about $70\%$ the size of the sliding
window, e.g., $27\times27$ pixels for the $38\times38$ sliding
window, and $269\times269$ pixels for $385\times385$ sliding window.

The performance of the GoogLeNet-based detector on the $1900\times
1900$ pixels image corresponding to test-zone-1 are shown in
Table~\ref{swindow} and the corresponding heatmap to each window
size are illustrated in Figure~\ref{heatmaps}.
The best accuracies, highest recall and F1-measure and high
precision, were obtained for a window size of $64\times 64$ pixels.
The time needed to perform the detection process using this window
size was $291$ minutes. This represents the execution time that
would be required for {\it Ziziphus lotus} shrub detection on any
new input image of the same dimensions, which is time
consuming to be used in larger regions or across the entire range
distribution of the species along the Mediterranean region. To
reduce the execution time we next explored the use of appropriate
pre-processing techniques.

\begin{table}[h!]
\centering \caption{ The GoogLeNet CNN-detection results for {\it
Ziziphus lotus} shrub mapping in Test-zone-1 with and without
data-augmentation, both under the sliding window approach, and using
pre-processing. Accuracies are expressed in terms of true positives
(TP), false positives (FP), false negatives (FN), precision, recall,
and F1 measure. The highest accuracies are highlighted in bold.}
\begin{tabular}{|l|l|l|l|l|l|l|}
\hline Detection model & TP  & FP  & FN  & precision & recall & F1\\
\hline \hline
CNNs  (test-zone-1)          &  &  &  &          &        &\\
+fine-tuning &  &  &  &          &        &\\
+sliding window & 63 &12 &9  &   84.00\%    & 87.50\%&
85.71\%\\\hline +
CNNs  (test-zone-1)          &  &  &  &          &        &\\
+fine-tuning &  &  &  &          &        &\\
+augmentation &  &  &  &          &        &\\
+sliding window & 65 & 7& 7 &  90.28\% & 90.28\% & 90.28\% \\\hline
CNNs  (test-zone-1)          &  &  &  &          &        &\\
+fine-tuning &  &  &  &          &        &\\
+augmentation &  &  &  &          &        &\\
+pre-processing     & 69& 1 & 3  &  {\bf 98.57\%} & {\bf 95.83\%} &{\bf
97.18\%}\\\hline
\end{tabular}
 \label{fresults1}
\end{table}

\subsubsection{Detection with image pre-processing}
\begin{figure}[h!]
\centering
\includegraphics[width=1.0\textwidth]{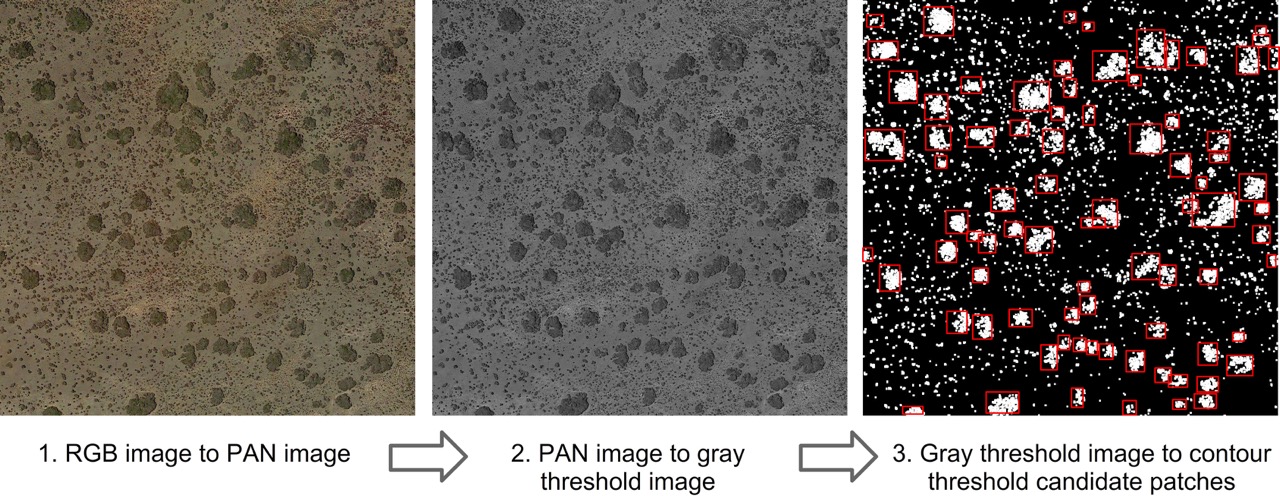}
\caption{ Pre-processing consisted in, first, converting the three
band image into one gray-scale band image (PAN), second, converting
the gray-scale image into a binary image based on a 100 over 256
digital value threshold, and third, detecting {\it Ziziphus lotus}
shrubs only in pixels with a digital value greater than 100. The 78
candidate patches identified in Test-zone-1 are labeled with red
contour in the right panel (the 53 candidates in Test-zone-2 are not
shown.} \label{Prep:PP}
\end{figure}

To optimize the CNN-detection accuracy and execution time, we
analyzed the impact of several pre-processing techniques. It is
worth to mention that the set of pre-processing techniques that
provides the best results depends on the nature of the problem and
the object of interest. From the multiple techniques explored, the
ones that improved the performance of the detector in this work
were: i) Eliminating the background using a threshold based on the
high albedo (light color) of the bare soil. For this, we first
converted the RGB image to gray scale and then created a binary
mask-band to select only those pixels darker than 100 over 256
digital levels of gray, which was the average level of gray of the
field georeferenced polygons of bare soil in test-zone-1.
ii) Applying an edge-detection method to the previously created mask-band
to select only clusters of pixels with an area greater than 180 pixels (21.6 m$^2$), which
approximately was the
size of the smallest {\it Ziziphus lotus} individual georeferenced in test-zone-1.
After pre-processing,
the number of candidate image patches to pass to the CNN detector was 78 for test-zone-1
and 53 for test-zone-2, what significantly decreased the detection computing time. See
illustration of the pre-processing phase in Figure~\ref{Prep:PP}.

The results of the CNN-based detection, considering all the
optimizations in the training and detection steps described above
for test-zone-1 are summarized in Table~\ref{fresults1}. The CNN
model reached relatively good accuracies ($>84\%$) using only fine-tuning
and very good accuracies when adding data-augmentation ($>90\%$) under
the sliding-window detection approach. Using pre-processing techniques
together with fine-tuning and data-augmentation clearly improved the
precision, recall and F1 by $14.57\%$, $8.33\%$
and $11.47\%$, respectively, compared with the baseline  CNN-model with fine-tuning and under the sliding window approach.

\begin{table}[h!]
\centering \caption{A comparison between GoogLeNet and OBIA-based
detectors,  on test-zone-2,  in terms of true positives (TP), false
positives (FP), false negatives (FN), precision , recall and
F1\_measure. The highest values are highlighted in bold.}
\begin{tabular}{|l|l|l|l|l|l|l|}
\hline Detection model & TP  & FP  & FN  & precision & recall & F1\\
\hline \hline
CNNs  (test-zone-1)          &  &  &  &          &        &\\
+fine-tuning &  &  &  &          &        &\\
+augmentation &  &  &  &          &        &\\
+pre-processing     & 69& 1 & 3  &  {\bf 98.57\%} & {\bf 95.83\%} &{\bf
97.18\%}\\\hline\hline
OBIA  (test-zone-1)          & 66 & 19& 6 & 77.65\%           & 91.67\%        &84.08\% \\
\hline \hline
CNNs  (test-zone-2)          &  &  &  &          &        &\\
+fine-tuning &  &  &  &          &        &\\
+augmentation &  &  &  &          &        &\\
+pre-processing     & 38& 3 & 3  & {\bf92.68\%}&     {\bf92.68\%}&{\bf
92.68\%}
\\\hline\hline
OBIA  (test-zone-2)           & 21 & 8& 20 & 72.41\%&51.00\%& 60.00\%\\
\hline
\end{tabular}
 \label{fresults2}
\end{table}

\subsection{CNN- versus  OBIA-detector}

\begin{figure}[h!]
\centering
\includegraphics[width=1.0\textwidth]{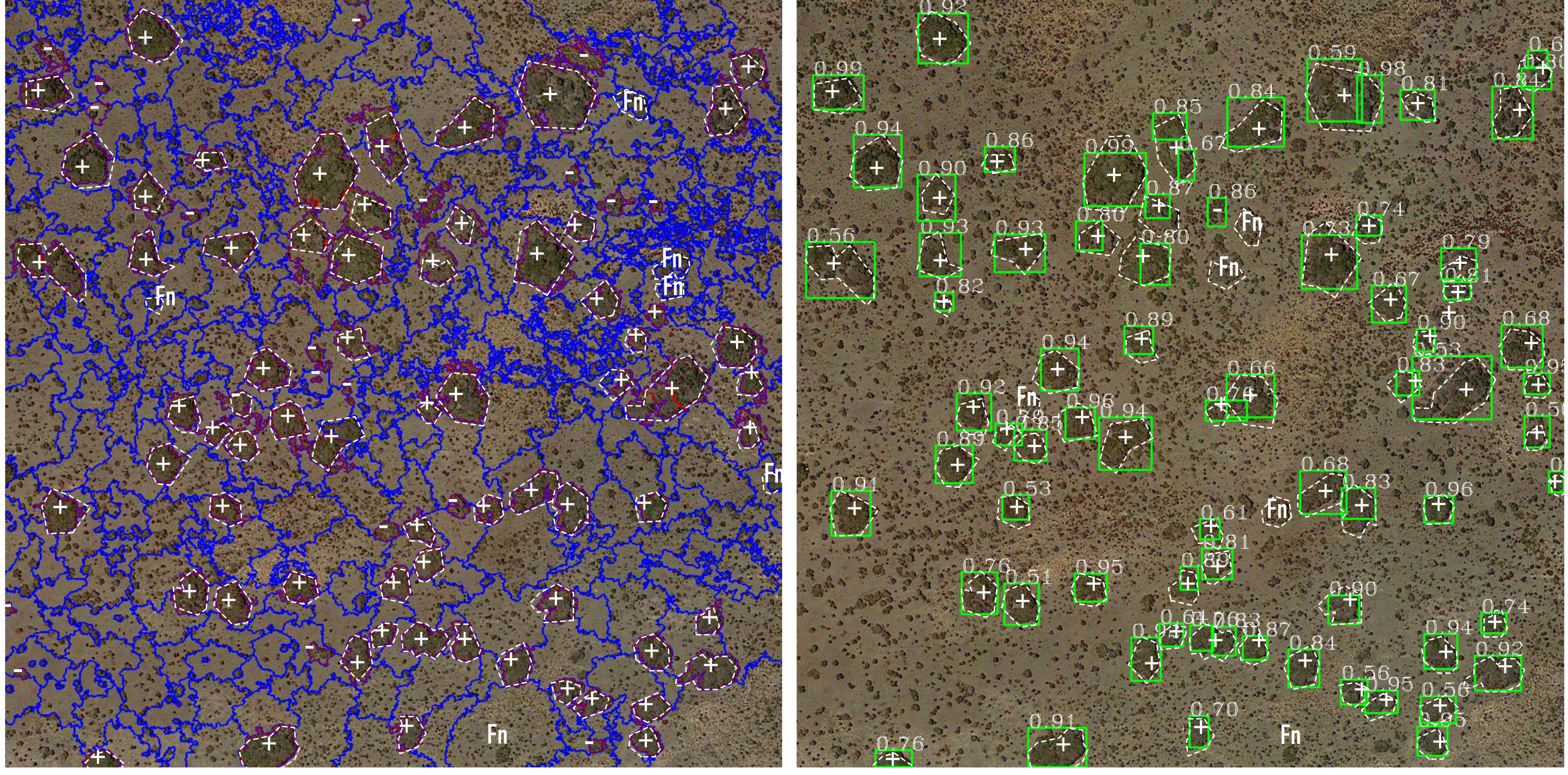}
(a) OBIA detection on test-zone-1~~~~(b)CNN detection on test-zone-1
\includegraphics[width=1.0\textwidth]{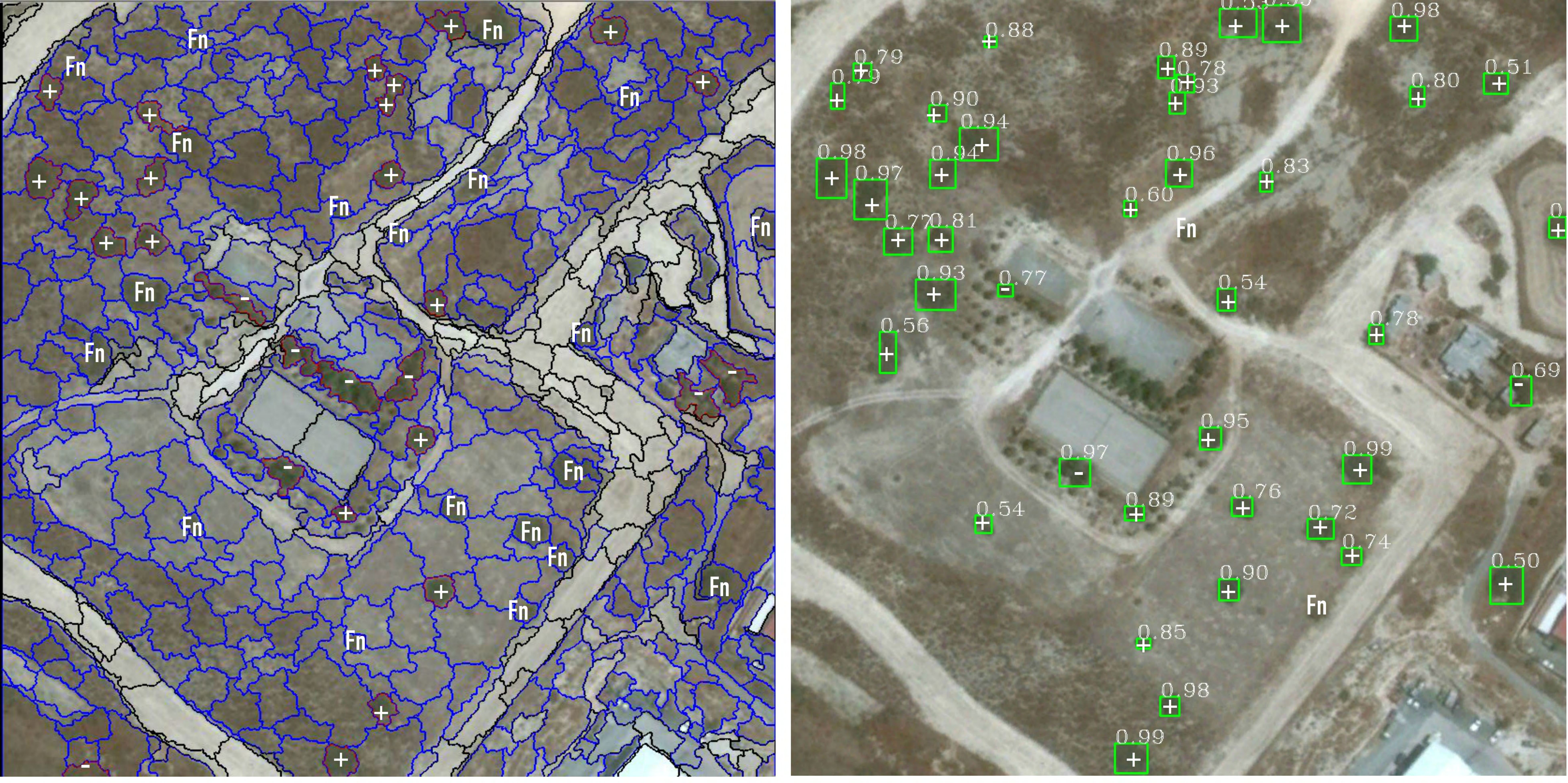}
(c) OBIA detection on test-zone-2~~~~(d) CNN detection on
test-zone-2 \caption{The detection results obtained by OBIA-based
model on test-zone-1 (a) and test-zone-2 (c)  and CNN-based model on
test-zone-1 (b) and test-zone-2 (d). The symbols (+), (-) and (Fn)
stand for true positive, false positive and false negative
respectively. The values on the top of the bounding boxes in (2) and
(4) show the probabilities, calculated by GoogLeNet-detector, of
having a {\it Ziziphus lotus} shrubs.} \label{Prep}
\end{figure}

To obtain the best detection results using OBIA-method, the user has
to manually determine the best segmentation parameters and
classification procedure. In this work, we iteratively tried all
possible combinations among parameters: scale ranging in $[80,160]$
at intervals of 5, shape and compactness ranging in $[0.1,0.9]$ at
intervals of 0.1. An exhaustive search of the best configuration
implied the evaluation of $2511$ combinations, each segmentation
test took $18$ seconds, and each classification test took around
$10$ seconds. The whole detection process using OBIA took around
$12.5$ hours. The best segmentation parameters were: scale = 110,
shape $= 0.3$, compactness $= 0.8$. The best classification
configuration was: Brightness $< 80.87$, Red Band $> 97.35$, Green
Band $< 81.18$, Blue Band $<64.96$ and gray level co-occurrence
matrix (GLCM mean) $< 80.54$. To ensure a fair comparison with CNNs,
we re-utilized the segmentation and classification configuration
from test-zone-1 in test-zone-2. It is important to recall that OBIA
requires the user to provide training points of each input class
located within the scene to classify.

To test the performance of the CNN classifier in a region with different characteristics
and far away from the training zone, we compared the performance of GoogLeNet-based model
and the OBIA-method in test-zone-1 in Spain and  test-zone-2 in Cyprus
(results summarized in Table~\ref{fresults1}). As we can observe from these tables,
CNN-based detection model achieved significantly better detection
results than OBIA on both test zones.  On test-zone-1, CNN achieves
higher precision, 98.57\% versus 77.65\%, higher recall, 95.83\%
versus 91.67\%, and higher F1-measure, 95.83\% versus 84.08\% than OBIA.
Similarly, on test-zone-2, CNN achieves significantly better
precision, recall  and F1-measure than OBIA. The detection results of CNN and OBIA, on
test-zone-1, are shown in Figure~\ref{Prep}(1) and (2) and on
test-zone-2 are shown in Figure~\ref{Prep}(3) and (4).

In terms of user productivity, the training of the CNN-classifier
with fine-tuning and data augmentation were performed only once and
took 7.55 minutes on two NVIDIA GeForce GTX 980
GPUs. In the  deploying phase, using GoogLeNet-classifier and pre-processing on
test-zone-1 and test-zone-2 took 35.4 and 24.1 seconds respectively. Whereas,  finding the best configuration for OBIA on each test-zone took around 12.5 hours. Applying the obtained CNN-detector to any new image of similar sizes will take seconds; however, applying OBIA to a new image will take 12.5 hours.
This clearly shows that  the user
becomes more productive with CNNs.

\section{Conclusions}

In this work, we explored, analyzed and compared two detection
methodologies, the OBIA-based approach and the CNNs-based approach,
in the challenge of mapping {\it Ziziphus lotus}, a shrub of
priority conservation interest in Europe. Our experiments
demonstrated that the GoogLeNet-based classifier with transfer
learning from  ImageNet and data augmentation, together with
pre-processing techniques provided better detection results than
OBIA-based methods. In addition, a very competitive advantage of the
CNN-based detector is that it required less human supervision than
OBIA and can be easily ported to other regions or scenes with
different characteristics, i.e., color, extent, light, seasons.

The proposed CNN-based approach can be systematized and reproduced
in a wide variety of object detection problems or land-cover mapping
using Google Earth images. For instance, our CNN-based approach
could support the detection and monitoring of trees and arborescent
shrubs, which has a huge relevance for biodiversity conservation and
carbon accounting worldwide. The presence of scattered trees have
been recently highlighted as keystone structures capable of
maintaining high levels of biodiversity and ecosystem services
provision in open areas~\cite{Prevedello2017}. Global initiatives
could greatly benefit from the CNNs, such as those recently
implemented by the United Nations Food and Agricultural
Organization~\cite{bastin2017extent} to estimate the overall
extension of forests in drylands biomes, where they used the
collaborative work of hundreds of people that visually explored
hundreds of VHR images available from Google Earth to detect the
presence of forests in drylands.

\section*{Acknowledgments}

We would like first to thank Ivan Poyatos and Diego Zapata for their
technical support. Siham Tabik was supported by the Ram\'on y Cajal
Programme (RYC-2015-18136). The work was partially supported by the
Spanish Ministry of Science and Technology under the projects:
TIN2014-57251-P, CGL2014-61610-EXP, CGL2010-22314 and grant
JC2015-00316, and ERDF and Andalusian Government under the projects:
GLOCHARID, RNM-7033, and P09-RNM-5048. This research was also
developed as part of project ECOPOTENTIAL, which received funding
from the European Union Horizon 2020 Research and Innovation
Programme under grant agreement No. 641762, and by the European LIFE
Project ADAPTAMED LIFE14 \newline
 CCA/ES/000612.

\bibliographystyle{plain}
\bibliography{sample}

\end{document}